\documentclass[letterpaper, 10 pt, conference]{ieeeconf}  

\IEEEoverridecommandlockouts                              

\usepackage{amsmath} 
\usepackage{amssymb}  
\usepackage{graphicx}
\usepackage{multirow}
\usepackage{makecell}
\usepackage{subcaption}
\usepackage{balance}
\usepackage{color}
\usepackage{hyperref}
\usepackage{enumerate}
\usepackage{times}
\usepackage{epsfig}
\usepackage{hhline}
\usepackage{mathtools, nccmath}
\usepackage{siunitx}
\usepackage{float}
\let\labelindent\relax
\usepackage{enumitem}
\usepackage[export]{adjustbox}

\usepackage{wrapfig,lipsum,booktabs}
\title{\LARGE \bf
Closing the Reality Gap: Zero-Shot Sim-to-Real Deployment for Dexterous Force-Based Grasping and Manipulation
}

\author{Zhe Zhao$^{1}$, Zhibin Li$^{2}$, Yilin Ou$^{1}$, and Mengshi Qi$^{1}$\thanks{Corresponding author: Mengshi Qi ({\tt\small qms@bupt.edu.cn}).}\\
$^{1}$State Key Laboratory of Networking and Switching Technology\\
Beijing University of Posts and Telecommunications, China\\
$^{2}$University College London, United Kingdom
}

\begin{document}

\maketitle
\thispagestyle{empty}
\pagestyle{empty}

\begin{abstract}
Human-like dexterous hands with multiple fingers offer human-level manipulation capabilities but remain difficult to train the control policies that can deploy on real hardware due to contact-rich physics and imperfect actuation. We present a sim-to-real reinforcement learning that leverages dense tactile feedback combined with joint torque sensing to explicitly regulate physical interactions. To enable effective sim-to-real transfer, we introduce (i) a computationally fast tactile simulation that computes distances between dense virtual tactile units and the object via parallel forward kinematics, providing high-rate, high-resolution touch signals needed by RL; (ii) a current-to-torque calibration that eliminates the need for torque sensors on dexterous hands by mapping motor current to joint torque; and (iii) actuator dynamics modeling with randomization to account for non-ideal torque–speed effects and bridge the actuation gaps. Using an asymmetric actor–critic PPO pipeline, we train policies entirely in simulation and deploy them directly to a five-finger hand. The resulting policies demonstrated two essential human-hand skills: (1) command-based controllable grasp force tracking and (2) reorientation of objects in the hand, both of which were robustly executed without fine-tuning on the robot. By combining tactile and torque in the observation space with scalable sensing/actuation modeling, our system provides a practical solution to achieve reliable dexterous manipulation. To our knowledge, this is the first demonstration of controllable grasping on a multi-finger dexterous hand trained entirely in simulation and transferred zero-shot on real hardware.

\end{abstract}

\section{INTRODUCTION}
Multi-finger dexterous hands offer human-like manipulation capabilities, but their high degrees of freedom and complex contact dynamics make control challenging. Most real-world applications still rely on simple parallel-jaw grippers, and robust \emph{in-hand manipulation} remains an open problem.

This work explores the use of \emph{full-state feedback}—tactile sensing and motor currents—to learn policies that explicitly sense and regulate contact forces. Unlike vision- or kinematics-only methods, we provide the policy with high-resolution tactile data and motor currents, capturing detailed interaction forces. Using these inputs, we train for two key skills: (i) \emph{force-adaptive grasping} that tracks commanded grip forces, and (ii) \emph{in-hand object rotation}.

Deep reinforcement learning has advanced dexterous manipulation, with methods such as large-scale domain randomization for sim-to-real transfer of object rotation~\cite{andrychowicz2018learning}. Subsequent work in automatic domain randomization on a Rubik’s Cube task~\cite{akkaya2019rubik} showed that sim-to-real is the key~\cite{tobin2017dr}. Recent simulation improved fidelity and robustness for cube reorientation on an Allegro Hand~\cite{handa2023dextreme}. Tactile sensing further enhances dexterity, e.g., via visuotactile fusion~\cite{yuan2023robotsynesthesia} or binary touch signals~\cite{yin2023rotating}. Related advances include dexterous grasping~\cite{zhang2025robustdexgrasp}, humanoid manipulation~\cite{lin2025humanoid}, and in-hand manipulation of thin objects with binary tactile sensing~\cite{hu2025dexterous}, showing that sim-to-real RL can produce transferable policies.

\begin{figure}[t]
    \centering
    \includegraphics[width=1\linewidth]{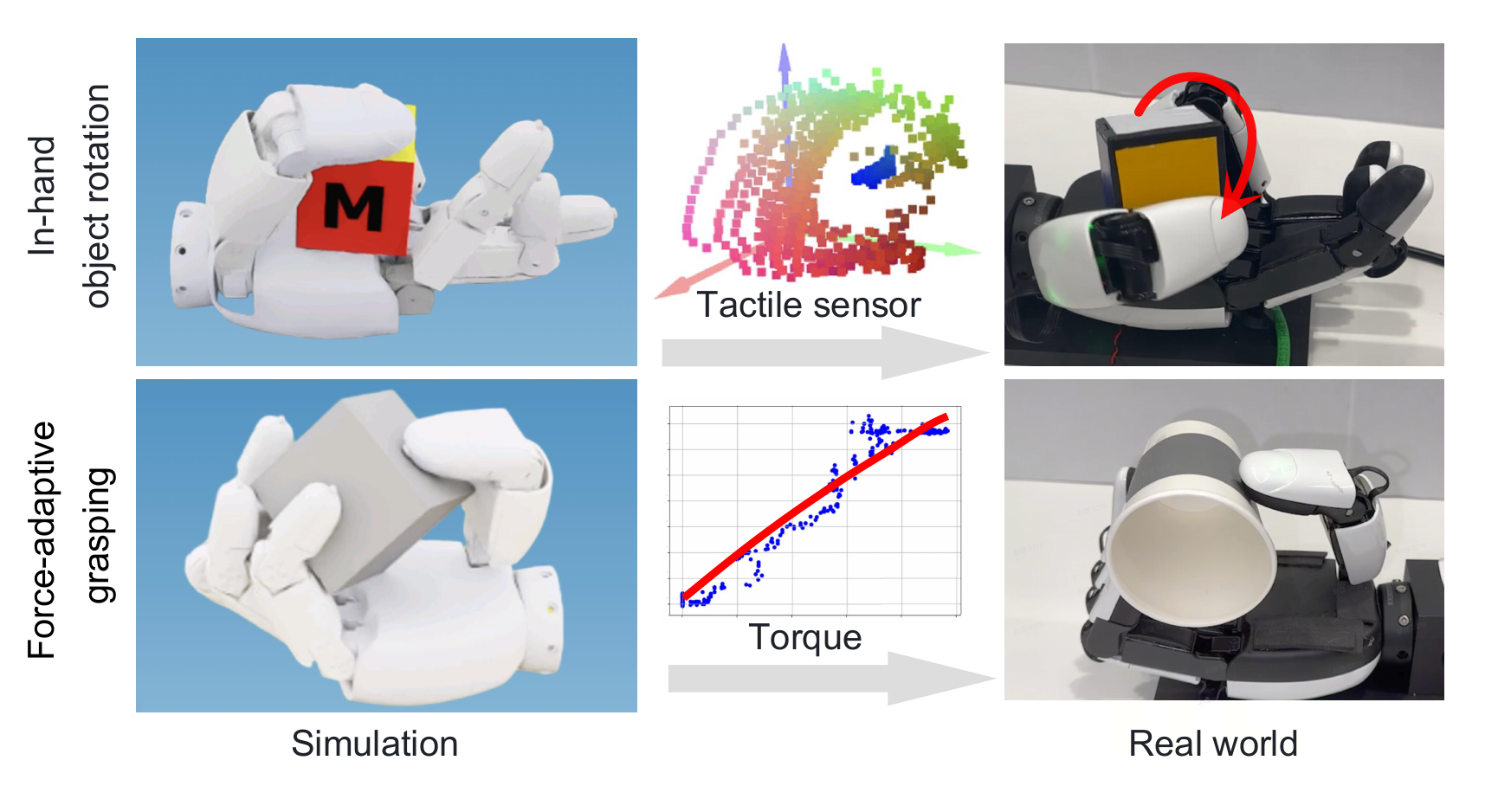}
    \caption{Learning full-state policy with tactile sensing and joint torques for dexterous grasping and in-hand manipulation.}
    \label{fig:placeholder}
    \vspace{-7mm}
\end{figure}

However, two practical hurdles limit the use of tactile sensing and joint torques: (i) \textit{Simulation of dense tactile contacts}: Simulating high-resolution tactile/contact is computationally too slow for massive RL exploration, resulting in a brittle policy due to insufficient exploration of contact dynamics. (ii) \textit{Discrepancies of actuator dynamics and modeling}: Most hands lack joint torque sensors and use current-based estimates; lack of capturing motor dynamics in simulation hinders the sim-to-real transfer.

As high-DoF dexterous hands become more available, we are motivated to develop a reliable sim-to-real \textit{recipe} for learning full-state policies with tactile and joint torque/current feedback that achieves zero-shot deployment. Our contributions include:
\begin{itemize}
[leftmargin=*]
    \setlength{\itemsep}{1pt}
    \setlength{\parskip}{0pt}
    \setlength{\parsep}{0pt}
\item \textbf{Full-state policies.} We formulate the policy observations that jointly take dense tactile sensing and joint torques, enabling explicit feedback that can infer contact states and contact forces.
\item \textbf{Fast, high-resolution tactile simulation.} We approximate tactile contacts by computing distances between a dense array of \emph{virtual tactile units} and the object via parallel forward kinematics, offering more accurate, high-rate tactile feedback suitable for massive RL training.
\item \textbf{Data-driven actuator modeling.} We bridge the sim-to-real actuation gap by calibrated current-to-torque mapping and randomization to account for uncertain motor dynamics (e.g., torque-velocity curves), requiring no direct torque sensors and reducing sim-to-real torque discrepancies on real hardware.
\item \textbf{Contact-rich sim-to-real RL policy training.} We trained and deployed two contact-rich skills: (1) \emph{controllable grasping} that tracks commanded grip forces and (2) \emph{in-hand object reorientation} -- full-state policies directly deployed for force-modulated grasps and reorientations on the real robot.
\end{itemize}



In summary, we contributed an easy-to-replicate \textit{recipe} for training full-state RL policies on a widely available 12-DoF dexterous hand. Our approach integrates tactile and motor-current sensing with computationally-fast simulation and robust actuator modeling, for zero-shot Sim-to-Real transfer of force-controllable grasping and in-hand manipulation. To our knowledge, this is the first demonstration of \emph{controllable grasping} with grip-force tracking, force-adaptive grasping, and robust in-hand manipulation on a multi-finger hand.


\section{Related Works}
\label{sec:related}

We review research work closely related to our focused techniques in our sim-to-real pipeline for dexterous hands, covering human-centric perception, grasp priors, tactile simulation, motor dynamics modeling, and current-to-torque estimation without direct joint torque sensors.

\subsection{Human-Centric Perception and Grasp Priors}
Understanding human actions and hand-object interactions provides useful context for dexterous manipulation, even when the final control problem is robotic. Recent multimodal perception methods have studied complementary cues for image/video segmentation~\cite{qi2026dcsam}, audiovisual commonsense reasoning~\cite{qi2026rdcl}, action form and quality assessment~\cite{qi2026efa,qi2025hpmcore}, balanced multimodal 3D human pose estimation~\cite{qi2026balancedhpe}, weakly-supervised temporal action localization~\cite{yun2024ssnf}, cross-modal video retrieval~\cite{qi2021sastb}, few-shot video classification~\cite{qi2020slowfast}, and sports video captioning~\cite{qi2020sportscaption}. These works show that heterogeneous visual, temporal, and semantic cues can improve human-centric understanding. In contrast, our focus is not high-level video understanding, but closed-loop dexterous control with tactile and motor-current feedback.

More directly related to hand-object interaction, temporal-consistent self-training has been used for 3D hand-object joint reconstruction under synthetic-to-real domain shift~\cite{qi2020tcsthandobject}. Human grasp generation methods have also modeled part-aware hand-object interactions with decomposed vector-quantized representations~\cite{zhao2024dvqvae} and extended this idea to both rigid and deformable objects~\cite{qi2025dvqvae2}. These reconstruction and generation studies provide useful human-like grasp priors from visual and generative modeling perspectives, whereas our work learns robot control policies for force-aware grasping and in-hand reorientation, and transfers them from simulation to real hardware without fine-tuning.

\subsection{Tactile Simulation}
High-fidelity, fast tactile simulation is a long-standing obstacle for learning contact-rich skills. For visuotactile sensors, recent systems libraries have pushed both realism and throughput. Akinola \emph{et al.} builds a library within Isaac Gym that synthesizes visuotactile images and contact-force distributions, and couples them with a policy-learning toolkit aimed at sim2real transfer~\cite{akinola2025tacsl}. In parallel, Nguyen \emph{et al.} integrate a soft-body FEM simulator with an optical visuotactile rendering pipeline inside Isaac Sim to capture elastomer deformation for GelSight-like skins~\cite{nguyen2024tacex}. Zhang \emph{et al.} focus on modeling multi-mode tactile imprints induced by different surface coatings/patterns and report high realism while remaining efficient enough for learning loops~\cite{zhang2025tacflex}. Beyond single-hand settings, works begin to exploit visuo–tactile simulation for complex, often bimanual, fine assembly: e.g., general sim2real protocols for marker-based visuotactile sensors~\cite{chen2024gpvt}, bimanual visuotactile assembly via simulation fine-tuning~\cite{huang2025vtrefine}.

On the policy side, tactile-only or visuo–tactile in-hand reorientation has seen rapid progress. Yin \emph{et al.}learns touch-only in-hand rotation on low-cost binary sensors, emphasizing robustness to sensing imperfections~\cite{yin2023rotating}. Yuan \emph{et al.} propose a point-cloud tactile representation to fuse vision and touch for in-hand rotation~\cite{yuan2024robotsynesthesia}. Purely tactile in-hand manipulation with a torque-controlled hand was shown in~\cite{sievers2022tactile}, while works also demonstrate DRL-based tactile control for slender cylindrical objects~\cite{hu2025slender}. For broader dexterous/bimanual touch, Lin \emph{et al.} studies bimanual tactile manipulation with sim-to-real deep RL~\cite{lin2023bitouch}. Collectively, these efforts motivate fast, high-fidelity tactile simulation and policy training; our distance-field–based tactile simulation targets this efficiency–fidelity trade-off specifically for RL.

\subsection{Motor Modeling}

For dexterous hands and manipulation, recent systems explicitly incorporate motor/drive limits or task-informed SysID. Huang \emph{et al.} employ privileged learning, SysID, and DR to transfer functional grasps across diverse hands~\cite{huang2024fungrasp}. Tactile in-hand works with torque-controlled hands to build policies on top of explicit joint torque models~\cite{sievers2022tactile,hu2025slender}. In non-prehensile manipulation, early sim-to-real studies showed the importance of accurate actuation/friction models plus ensemble dynamics to combat identification error~\cite{lowrey2018nonprehensile}.  

In our setting, we explicitly model the motor’s \emph{torque–speed} envelope~\cite{shin2023actuatorconstrainedreinforcementlearninghighspeed}, then \emph{randomize} these parameters (stall torque, speed constants, friction/ripple surrogates) to cover manufacturing tolerances and temperature/load variation. This follows the spirit of recent agile-mobility works where actuator models are aligned via residual learning or unsupervised actuator nets~\cite{he2025asap,fey2025uan}, but adapts them to the brushed-DC, gear-driven fingers typical of dexterous hands.

\subsection{Current-to-Torque Calibration}
Most dexterous hands available nowadays typically report \emph{motor current} but lack torque sensing. When torque feedback (or torque-conditioned policy inputs) is desired, estimating a reliable \emph{current$\to$torque} mapping becomes critical. In industrial/manipulator literature, motor-current–based wrench estimation has a long history: Kalman/filtering and momentum-observer formulations estimate Cartesian forces/torques from joint currents and states~\cite{wahrburg2017tase,han2022sensorless}. 
Gold \emph{et al.} consider torsional deflection plus motor current to estimate joint torques~\cite{gold2019ifac}. 
From a calibration/control perspective, \emph{current-based impedance control} explicitly fits actuator current/torque ratios and friction for current-controlled robots—eschewing force/torque sensors~\cite{dewolde2024currentimpedance}. Identification methods that retrieve physically consistent drive gains and dynamics are also relevant~\cite{gautier2014drivegains}.  

Compared to these, we target \emph{brushed-DC} finger actuators in dexterous hands. We fit a current$\to$torque map under quasi-static conditions to capture effective torque constants and biases (including gear train losses), 
Empirically, this provides torque observables for policies and improves sim–real alignment without adding expensive joint-torque sensors.



\section{Methodology}
\label{sec:method}
This section first presents the problem formulation, i.e., controllable grasping and in-hand object reorientation. Then, we present the technical details of our full-state policy learning framework, including the design of the full-state observation and the recipe of sim-to-real transfer.

\subsection{Problem formulation}
This work addresses the \textbf{sim-to-real transfer} of dexterous manipulation policies for multi-fingered robotic hands, focusing on zero-shot deployment of simulation-trained policies onto real hardware. Despite progress in RL-based manipulation, transferring contact-rich, force-sensitive tasks—such as variable-force grasping and in-hand reorientation—remains challenging due to gaps in tactile simulation, actuator modeling, and contact physics.

Formally, given a dexterous hand with $N$ fingers equipped with tactile sensors and current-controlled motors (no torque sensors), we learn a policy $\pi:o_t\mapsto a_t$ that maps observations in simulation that performs successfully on the physical system without fine-tuning.

\subsubsection{Force-Adaptive Grasping} Achieve stable grasps with specified force levels across objects of varying shapes and physical properties.

\subsubsection{In-Hand Object Rotation} Precisely rotate a grasped object around a defined axis through continuous, contact-rich multi-finger coordination.

Experiments use a 12-DoF direct-drive dexterous hand (XHand)~\cite{robotera2024generic} and are conducted in the IsaacLab simulation environment~\cite{mittal2023orbit}.

\subsection{Identification of Sim-to-Real Gaps}
Our study identified 4 primary sim-to-real gaps:

\textit{1) Perceptual Gap:} Simulations use ground-truth states, while real systems rely on noisy visual sensing (e.g., cameras), introducing errors from calibration and lighting.

\textit{2) Discrepancies in Actuator Dynamics:} Unmodeled joint-level discrepancies such as transmission backlash, nonlinear motor control, and response delays.

\textit{3) Discrepancies in Contact Physics:} Inaccurate contact modeling due to simplified surface geometry, material properties (e.g., stiffness, friction), and restitution.

\textit{4) Unseen State Distribution Shift:} Mismatch between training and real-world observations during deployment.

\subsection{Task Setup}
\subsubsection{Force-adaptive grasping} 
Traditional dexterous grasping methods imitate human scenarios: objects are placed on a table, and the hand starts from a random pose to attempt grasping. This approach requires extensive exploration, is computationally expensive, and often leads to unnatural grasps that fail in real environments due to sim-to-real gaps. To address these issues, we introduce a creative setup of inverted ``catch-the-object'', where the hand is fixed upside down and objects are dropped toward the palm (see Fig.~\ref{fig:grasp} in the later section). The policy learns to catch and stably grasp objects under randomized object properties, dynamics, and external perturbations--which simplifies the learning process, reduces complexity of reward exploitation, and produces robust grasping behaviors using only real-world measurable observations, facilitating better sim-to-real transfer.

\subsubsection{In-Hand Object Rotation}
For the object rotation task, a cube is rotated along a single axis due to the hand's limited degrees of freedom. Each successful 90-degree rotation about this axis triggers an update of the target pose by another 90°, enabling continuous rotation.

\subsection{Policy learning in simulation}

\begin{table}[]
\vspace{2mm}
\caption{Observation for force-adaptive grasping}
\vspace{-1mm}
\centering
\begin{tabular}{@{}c|ccc@{}}
\toprule
Input                  & Dimensionality & Actor                     & Critic                    \\ \midrule
Hand joints angles     & 12D            & \checkmark & \checkmark \\
Hand joints torque     & 12D            & \checkmark & \checkmark \\
Object position        & 3D             & \checkmark & \checkmark \\
Object linear velocity & 3D             & \checkmark & \checkmark \\
Contact force          & 5D             & \checkmark & \checkmark \\
Contact center         & 15D            & \checkmark & \checkmark \\
Fingertip positions    & 15D            & \checkmark & \checkmark \\
Force command          & 1D             & \checkmark & \checkmark \\ \bottomrule
\end{tabular}

\label{tab:RewardGrasping}
\vspace{-3mm}
\end{table}

\begin{table}[]
\caption{Observation for in-hand object rotation}
\vspace{-1mm}
\centering
\begin{tabular}{@{}c|ccc@{}}
\toprule
Input                       & Dimensionality & Actor & Critic \\ \midrule
Hand joints angles          & 12D            &  \checkmark     &  \checkmark     \\
Relative target orientation & 6D             &  \checkmark     &  \checkmark      \\
Last actions                & 12D            &  \checkmark     &  \checkmark      \\
Contact center              & 15D            &  \checkmark     &  \checkmark      \\
Contact force               & 5D             &  \checkmark     &  \checkmark      \\
Fingertip positions         & 15D            &  \checkmark     &  \checkmark      \\
Object orientation          & 6D             & ×     &  \checkmark      \\
Fingertip velocities        & 30D            & ×     &  \checkmark      \\
Fingertip rotations         & 30D            & ×     &  \checkmark      \\
Hand joints velocities      & 12D            & ×     &  \checkmark      \\
Object linear velocity      & 3D             & ×     &  \checkmark      \\
Object angular velocity     & 3D             & ×     &  \checkmark      \\ \bottomrule
\end{tabular}

\vspace{-5mm}
\label{tab:obs_rotate}
\end{table}

\subsubsection{Force-adaptive grasping} 
The observations are summarized in the Tab.~\ref{tab:RewardGrasping}. For the real-world setup, the object's Z-coordinate is estimated using the distance between the center of the object mask—inferred via SAM~\cite{kirillov2023segment}—and the palm, while the X and Y coordinates remain fixed. The object’s linear velocity is derived based on this positional information. The force command, which ranges from 0 to 1, is provided to the policy to indicate the desired grasping intensity.
During the simulation training process, we designed a reward mechanism to guide the policy toward achieving stable and reliable grasping during inverted catching tasks.

Once the fingers are in contact with the object, we employ torque command rewards $R_{\text{torque}}$ to encourage the agent to apply specific joint torques. The target torque $\tau_{\text{target}}$ is a continuous value provided by the instruction of the environment, and the reward function has a special handling for the thumb compared to the other fingers.

The total reward for this component is the sum of the individual finger torque rewards, which is applicable only when the finger is in contact with the object.
\begin{equation}
R_{\text{torque}} = w_{\text{torque}} \cdot \sum_{i \in \text{fingers}} R_{\text{torque}, i} \cdot I_{\text{contact}, i},
\end{equation}where $w_{\text{torque}}$ is the weight of this reward, $I_{\text{contact}, i}$ is the indicator function, which is 1 if the finger $i$ is in contact with the object and 0 otherwise.

The thumb has an opposing function, requiring greater force and a wider range of motion. Therefore, the reward for the thumb is binary, contingent on its torque $R_{\text{torque, thumb}}$ being within a valid range:
\begin{equation}
R_{\text{torque, thumb}} = I(0.01 \le \tau_{\text{thumb}} \le 1.1),
\end{equation}

Other four fingers' reward is a Gaussian function centered around the instructed target, multiplied by a validity mask.
\begin{equation}
R_{\text{torque}, i} = \exp\left(-\frac{(\tau_i - \tau_{\text{target}})^2}{2\sigma^2}\right) \cdot I(0.01 \le \tau_i \le 1.1) ,\quad
\end{equation}
\begin{figure}
\vspace{2mm}
    \centering
    \includegraphics[width=1\linewidth]{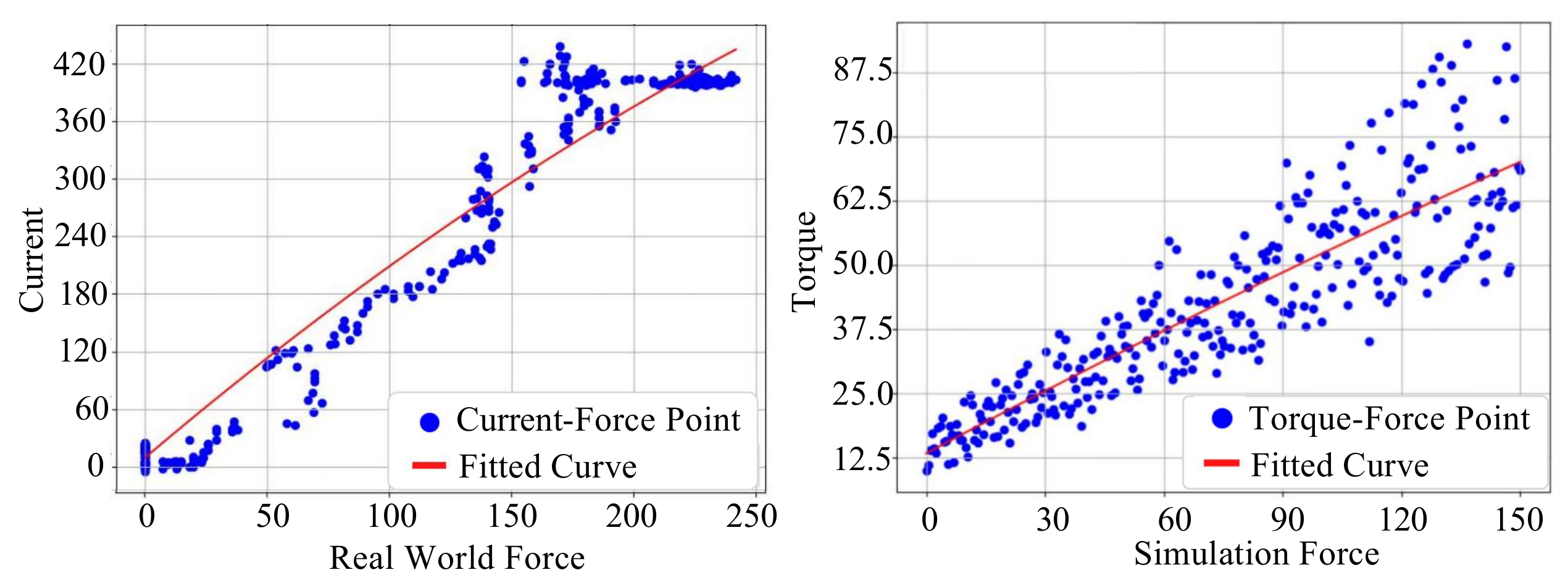}
    \caption{Calibration and alignment of \textit{current-force} (real robot) versus \textit{torque-force} (simulation) properties.}
    \label{fig:fig_force}
    \vspace{-5mm}
\end{figure}
For the contact force, similar to the joint torque reward, an appropriate contact force is encouraged.
\begin{equation}
R_{\text{force}} = w_{\text{force}} \cdot \sum_{i \in \text{fingers}} R_{\text{force}, i} \cdot I_{\text{contact}, i},
\end{equation}
\begin{equation}
    R_{\text{force, thumb}} = I(0.01 \le F_{\text{thumb}} \le 200),
\end{equation}

\begin{equation}
R_{\text{force}, i} = \exp\left(-\frac{(F_i - F_{\text{target}})^2}{2\sigma^2}\right) \cdot I(0.01 \le F_i \le 200) ,\quad
\end{equation}
where $F_i$ is the fingertip contact force of the root joint (finger $i$), $\sigma$ is the standard deviation of the Gaussian function, set at 0.10, $I(\cdot)$ is the indicator function.

In addition to standard grasping rewards, we introduce a novel four-finger consistency reward to promote coordinated motion among the four homologous and structurally similar fingers. This term regulates the uniformity of their flexion and extension angles. In the human hand, biomechanical constraints limit the range of independent motion of the metacarpophalangeal joints, causing adjacent fingers to move correlatively. In contrast, robotic fingers are driven by independent motors. Without explicit coordination, reinforcement learning policies often produce unnatural and inefficient postures. The consistency reward encourages more human-like motion patterns, leading to more stable and physically plausible grasps with improved force distribution.

The four-finger difference penalty $R_{\text{diff}}$ is defined as:
\begin{equation}
  R_{\text{diff}} = w_{\text{diff}} \cdot \text{Var}(\{q_j \mid j \in \text{inner fingers}\})  ,
\end{equation}
where $w_{\text{diff}}$ is the penalty weight, $q_j$ is position of the first joint of finger j, $\text{Var}(\cdot)$ is variance of the set of joint positions.

To penalize the deviation of the outer finger joints (index, middle, ring, little) from their central position, which causes ineffective grasping, the penalty term $R_{\text{outter}}$ is defined as:
\begin{equation}
    R_{\text{outter}} = w_{\text{outter}} \cdot \|{q_{\text{outter}}} - {c_{\text{outter}}}\|_2,
\end{equation}
where $w_{\text{outter}}$ is the weight of this penalty, ${q_{\text{outter}}}$ is current joint position vector for the outer joints, ${c_{\text{outter}}}$ is center position vector of the outer joints.

Also, several standard penalty terms are incorporated to refine the agent's behavior. A terminal state penalty is applied to prevent premature termination of the task. Furthermore, we penalize significant changes in action commands between consecutive time steps to encourage smoother control policies. This is formulated as the action rate penalty:
\begin{equation}
R_{\text{action}} = w_{\text{action}} \cdot \|{a}_t - {a}_{t-1}\|_2^2,
\end{equation}
where $R_{\text{action}}$ is the action rate penalty, $w_{\text{action}}$ is the weight of this penalty, ${a}_t$ is the action vector at the current time step $t$, and ${a}_{t-1}$ is the action vector at the previous time step $t-1$.
And a joint velocity L2 penalty to discourage high joint velocities. This encourages the agent to generate smoother and more stable movements, avoiding abrupt motions. The penalty is defined as:
\begin{equation}
R_{\text{vel}} = w_{\text{vel}} \cdot \|\dot{{q}}\|_2^2,
\end{equation}
where $\dot{{q}}$ represents the joint velocity vector, and $w_{\text{vel}}$ is the corresponding penalty weight.

\subsubsection{In-Hand Object Rotation}

\begin{table}[]
\vspace{2mm}
\caption{Reward for in-hand object rotation}
\centering
\begin{tabular}{@{}c|c|c@{}}
\toprule
Reward & Formula & Weight \\ \midrule
Close to goal & $ \frac{1.0}{|d_{rot}| + \epsilon}\times \frac{1}{1 + e^{400 (d_{goal} - 0.05)}}$ & 1.0 \\
Action penalty & $||{{a}_t - {a}_{t-1}}||^2$ & -0.0002 \\ \midrule
Reset reward & Condition & Value \\ \midrule
Reach goal bonus & $d_{rot}<0.3$ and $d_{pos}<0.05$ & 250 \\ \bottomrule
\end{tabular}

\vspace{-5mm}
\label{tab:cube_reward}
\end{table}

To mitigate the sim-to-real gap for rotating objects, we utilize observations that are commonly available in the real world, as shown in Tab.~\ref{tab:obs_rotate}. We implemented and successfully trained two types of policies: one without the direct measurement of object orientation, and the other with the feedback of the object orientation measured via an embedded IMU directly. 

The first type of policy excludes the relative target orientation from the observations and relies solely on the proprioceptive information of the dexterous hand to rotate the object. The second type of policy uses a low-cost IMU sensor embedded inside the object to obtain its orientation directly, and computes the relative rotation to the target. In particular, the technical know-how is to represent the orientation in a 6D continuous format~\cite{zhou2019continuity}, because the quaternion representation has a fundamental double-cover problem: every 3D rotation can be represented by two unit quaternions, q and -q, which correspond to the exact same orientation. This mathematical property creates an undesired discontinuity in the parameter space, disrupting policy learning. In contrast, the 6D rotation representation provides a continuous and smooth mapping, which in turn makes the loss function continuous and relatively smooth in the 6D parameter space.
To encourage continuous rotation, the agent is rewarded for returning the object to a reference position after rotation. However, unconstrained position error $R_{dist}$ initially dominates and creates a local optimum where the agent avoids rotation to preserve reward. To mitigate this, we combine position and rotation error into a single composite reward (Tab.~\ref{tab:cube_reward}), preventing such suboptimal behavior and promoting continuous in-hand manipulation.

\subsection{Sim-to-Real Transfer}

\begin{figure}
\vspace{2mm}
    \centering
    \begin{subfigure}[b]{0.48\linewidth}
        \centering
        \includegraphics[width=\linewidth]{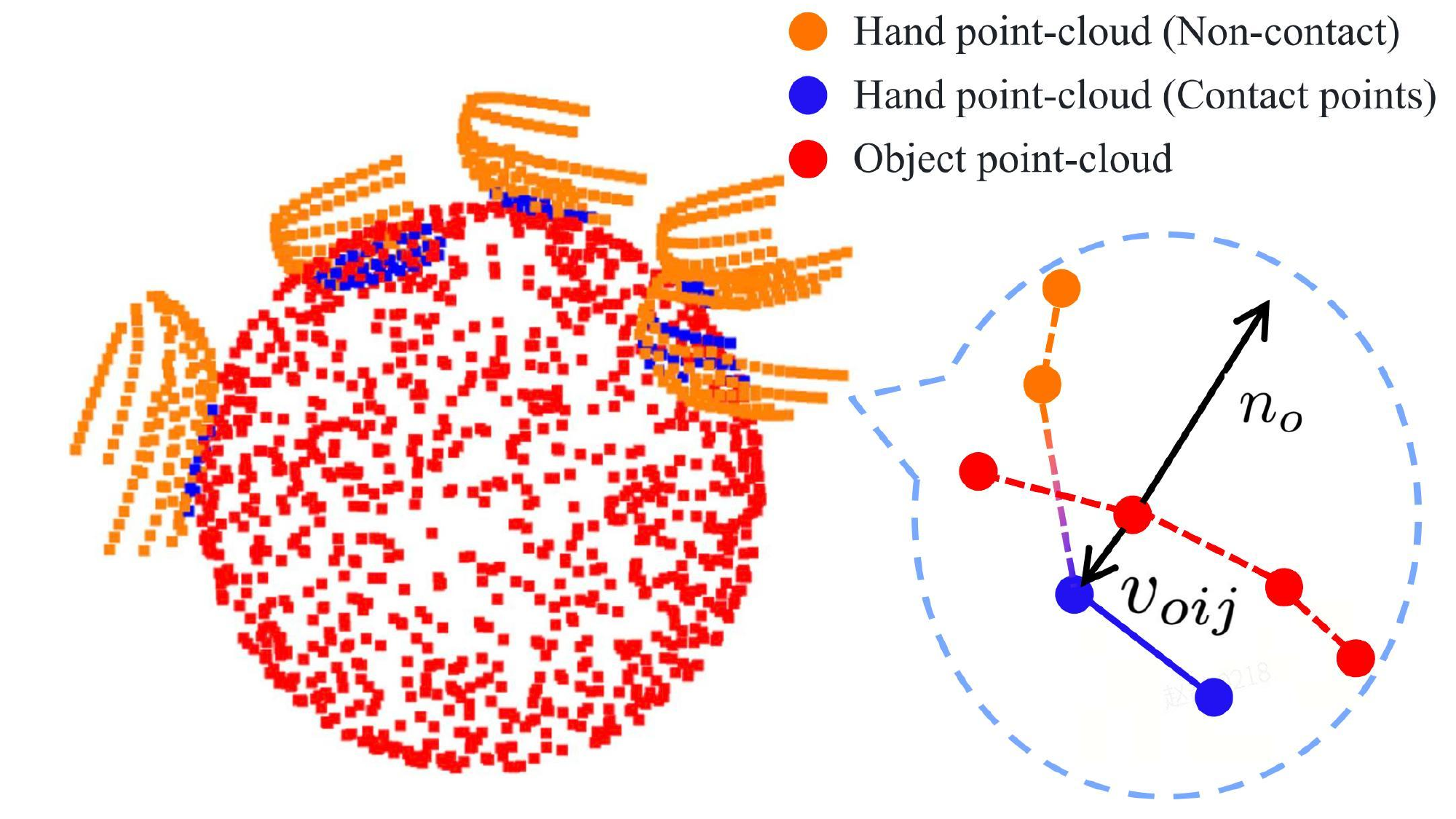}
        \caption{Modeling of contact points to approximate tactile sensors that mitigate the sim-to-real gap.}
        \label{fig:contact}
    \end{subfigure}
    \hfill
    \begin{subfigure}[b]{0.48\linewidth}
        \centering
        \includegraphics[width=0.9\linewidth]{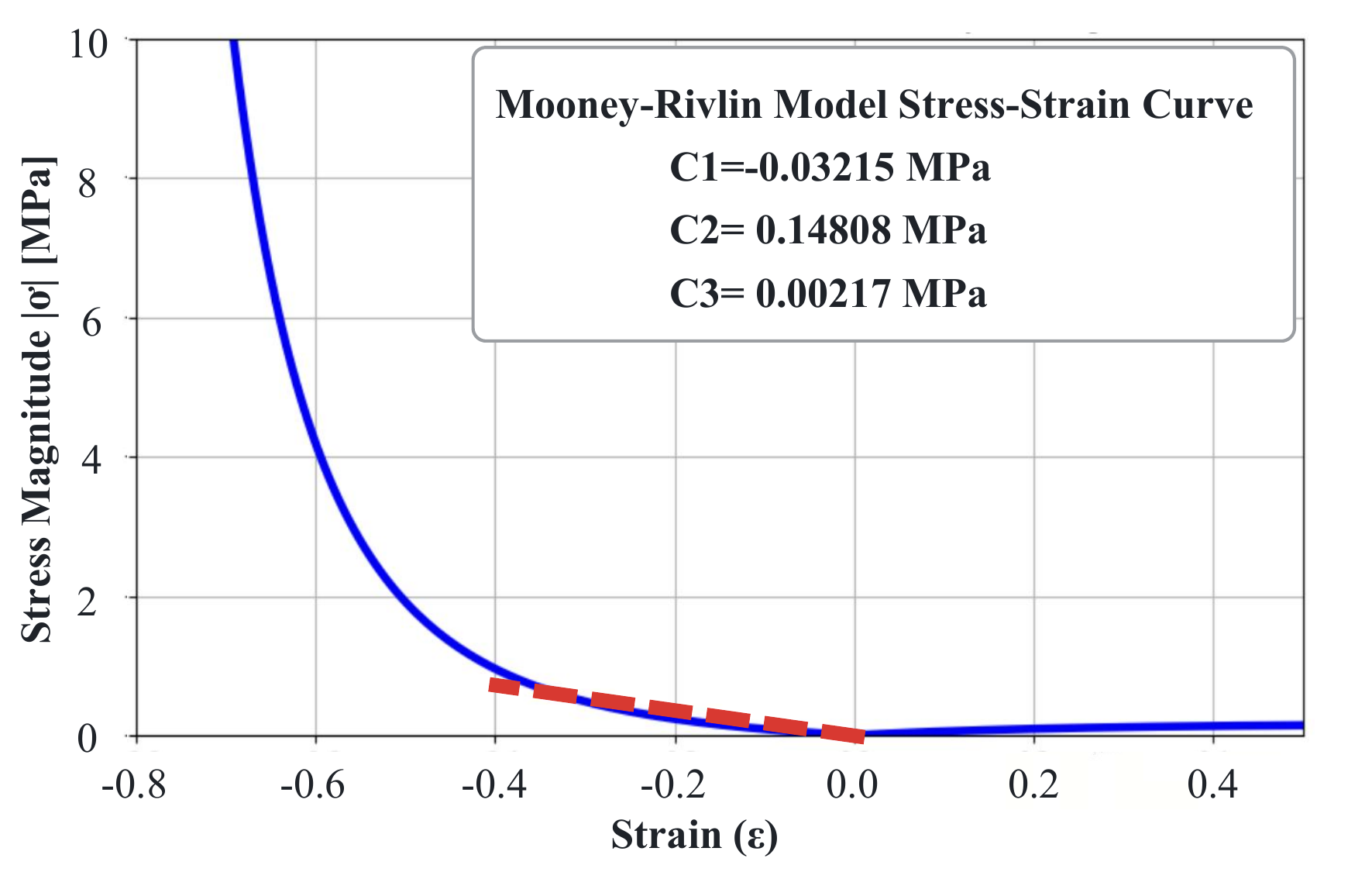}
        \caption{Stress-Strain curve of the rubber materials of the tactile sensor on the dexterous hands.} 
        \label{fig:Stress-Strain}
    \end{subfigure}
    \caption{Contact point modeling and material properties.}
    \label{fig:combined}
    \vspace{-6mm}
\end{figure}

We identified three primary contributors to the sim-to-real gap in our task: tactile sensing, complex in-hand contact simulation, and actuator dynamics. Our approach closes this gap through precise problem formulation and tailored simulation techniques.

\subsubsection{Tactile Sensor Simulation}

We implement a rapid, massively tactile sensor simulation using parallel forward kinematics, which calculates the distance from each sensor unit to the N-nearest surface points on the object.

We consider a dexterous hand with \(N\) fingers. Typically, \(N = 5\). The sensor set of the entire dexterous hand can be denoted as \(S\). On the fingertip of each finger \(i\) (where \(i \in \{1, 2, \ldots, N\}\)), there is a set of sensors, which we represent as \(S_i\). Therefore, the total sensor set is the union of all fingertip sensor sets:
\[ S = \bigcup_{i = 1}^{N} S_i. \]
For each individual sensor \(j\) on each fingertip \(i\) (where \(j \in \{1, 2, \ldots, M_i\}\), and \(M_i\) is the total number of sensors on the \(i\) - th fingertip), we define its key attributes:

\textbf{Position}: Through forward kinematics, we can calculate the spatial coordinates of each sensor from the hand degrees-of-freedom position, denoted as \(p_{ij} \in \mathbb{R}^3\).

\textbf{Activation}: On the real robot hand, the sensor is activated while reading is greater than a threshold. In the simulation, as shown in Fig.~\ref{fig:contact}, the nearest point ${p}_o$ on the surface of the object is found for each sensor ${p}_{ij}$. A vector from the object pointing to each \textit{tacxiel}, ${v}_{oij}$, is defined. The contact is detected if the sensor has deformed, which is determined by evaluating the dot product of ${v}_{oij}$ and the normal vector of the object at ${p}_o$. The vector from ${p}_o$ to ${p}_ij$ is given by:${v}_{oij} = {p}_{ij} - {p}_o$. Let ${n}_o$ be the outward-pointing normal vector at the object point ${p}_o$. If the dot product of ${v}_{oij}$ and ${n}_o$ is negative, it implies that the sensor has penetrated the object's surface, indicating a deformation. All points meeting such a condition are considered as \textit{equivalent contact points}.
    \begin{equation}
         \text{Contact} \iff {v}_{oij} \cdot {n}_o < 0 .
    \end{equation}

\textbf{Force}: On the real robot, normal forces are measured by the sensors. In the simulation, it is approximated using Mooney-Rivlin stress: strain relationship model to quantify the contact force of each sensor based on the distance $d_{ij} =|v_oij|$ between the sensor and the object surface and the total contact force per finger, denoted as \(f_{ij} \in \mathbb{R}^+\). 

In our model, we use the fully actuated robotic hand Xhand with 12 degrees of freedom. Each fingertip has \(M_i = 120\) sensors, so the entire hand has \(\sum_{i = 1}^{N}M_i = 600\) sensors. These sensors are concentrated in five independent ``dense groups''. This high local resolution on each fingertip can capture fine pressure distributions.

\subsubsection{Complex Contact Simulation}

The simulation of intricate in-hand contacts is made more robust by randomizing the coefficients for surface friction, damping, and restitution.

\textbf{Modeling of Tactile Sensors:} Building on the dual characteristics of contact sparsity and high resolution, the tactile information from each fingertip is denoted as $T_i$ to represent and process the tactile data. This modeling approach preserves critical contact states while accommodating the spatial sparsity of the data. $T_i$ is defined as:
\begin{equation}
    T_i = (F_i, {\mu}_i).
\end{equation}
where $F_i$ is the total contact force, ${\mu}_i$ is the Force-weighted Mean Contact Position.

\textbf{Total Contact Force ($F_i$):}
The resultant contact force $F_i$ is defined on the real robot as the sum of the total forces from all active sensors on the $i$-th fingertip. In simulation, it is the resultant force between the fingertip and the object, reflecting the effect of all the finger-object point-contacts:
\begin{equation}
    F_i = \sum_{j \in S_i^*} f_{ij}.
\end{equation}

\begin{figure}[t]
    \centering
    \vspace{2mm}
    \includegraphics[width=0.9\linewidth]{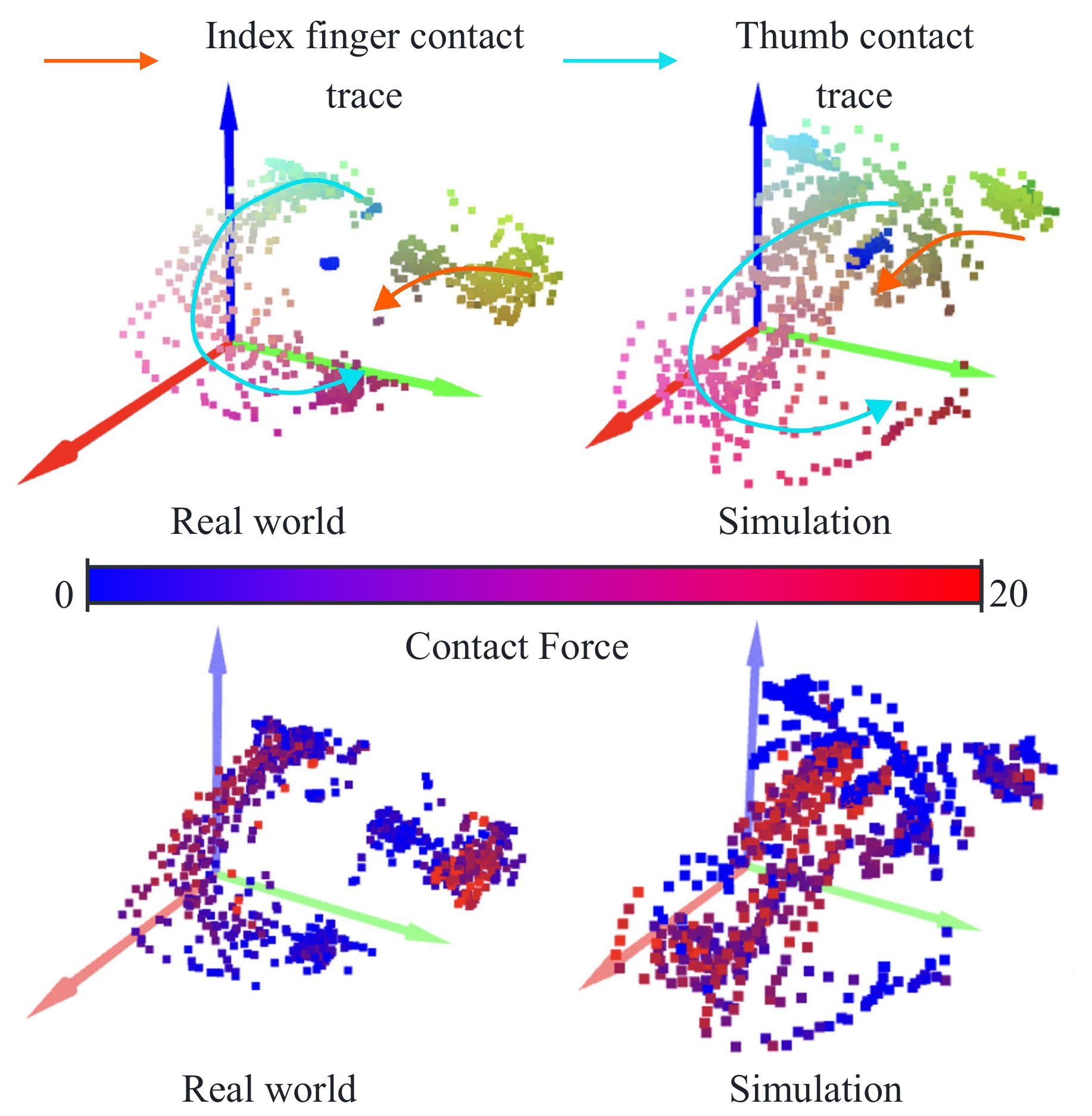}
    \caption{Visualization of real-world and simulated contact data during an in-hand rotation task. The close alignment between the contact points (Top) and contact forces (Bottom) shows the high fidelity of our contact simulation.}
    \label{fig:contact_point}
    \vspace{-5mm}
\end{figure}

\begin{figure*}[!thbp]
    \centering
    \includegraphics[width=1\linewidth]{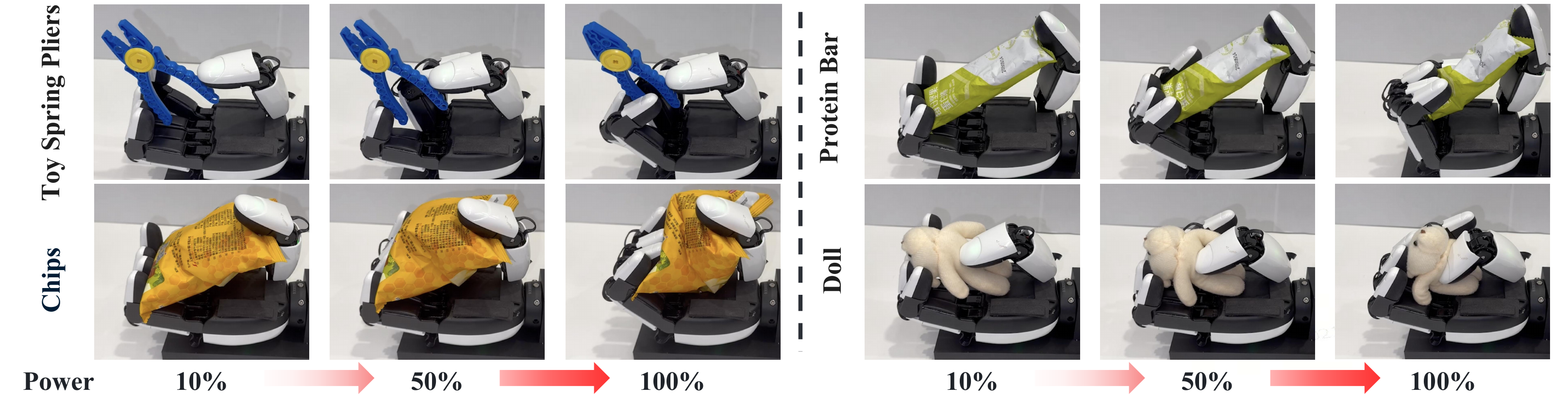}
    \caption{Grasping objects with controllable magnitudes of grasping forces -- from low to high strength.}
    \label{fig:grasp_force_command}
    \vspace{-5mm}
\end{figure*}

\textbf{Force-weighted Mean Contact Position (${\mu}_i$):}
The mean contact position ${\mu}_i \in \mathbb{R}^3$ is a force-weighted average position that represents the center of pressure:
\begin{equation}
    {\mu}_i = \frac{\sum_{j \in S_i^*} f_{ij} \cdot {p}_{ij}}{F_i}.
\end{equation}
Here, the position of each active sensor, ${p}_{ij}$, is weighted by its force value, $f_{ij}$. The denominator, $F_i$ (the total contact force), ensures this is a normalized weighted average. In a simulation environment, this process can be simplified to:

\begin{align}
      {\mu}_{i,sim} &=  \frac{\sum_{j \in S_i^*} d_{ij}\cdot k \cdot {p}_{ij}}{D_i\cdot k}\nonumber =\frac{\sum_{j \in S_i^*} d_{ij}\cdot {p}_{ij}}{D_i} .
\end{align}

As illustrated in Fig.~\ref{fig:contact_point}, we plot force-weighted contact centers and force magnitudes during an object selection task performed by a dexterous hand in both real-world and simulation environments. It can be observed that the distributions of force-weighted contact centers and force magnitudes exhibit high similarity between the real world and simulation. Furthermore, the real-world contact region is contained within the simulation contact region, demonstrating that contact features can be effectively transferred from simulation to the real world.
\subsubsection{Current-torque Calibration}
To calibrate real-world motor current and simulated joint torque with contact force, we applied varying forces to each fingertip in both real and simulated environments. Real-world motor current and tactile sensor readings were recorded, while joint torque and applied force magnitudes were collected in simulation. Scatter plots and fitted curves (Fig.~\ref{fig:fig_force}) show a proportional relationship between contact force and both current and torque. We recorded the maximum current $I_{max}$, maximum torque $\tau_{max}$, and their corresponding maximum contact forces $F_{max,real}$ and $F_{max,sim}$.  These values were normalized to $[0, 1]$, aligning real-world current with simulated torque and their respective contact forces.

\subsubsection{Actuator Model Randomization}


To bridge the sim-to-real gap in dexterous manipulation, we use a randomized actuator model that captures real-world motor nonlinearities—such as torque-velocity saturation and backlash hysteresis—improving policy robustness against hardware variations. The torque is computed via a PD controller with randomized gains:
\begin{equation}
\tau_c = k_p \cdot \left( q_{\text{ref}} - q_m \right) + k_d \cdot \left( \dot{q}_{\text{ref}} - \dot{q}_m \right),
\end{equation}
where $k_p$ and $k_d$ are the proportional and derivative gains, respectively, and $q_m$ denotes the measured joint position.

A key feature of the model is the incorporation of backlash, simulating mechanical dead zones due to gear play. The effective torque is modulated by a deadband function:
\begin{equation}
\tau_b = 
\begin{cases} 
0 & \text{if } |q_{\text{ref}} - q_m| < \epsilon \\
\tau_c & \text{otherwise}
\end{cases},
\end{equation}
where $\epsilon$ is the backlash threshold, randomized during training to mimic physical wear and tolerance variations.

Further, the torque is constrained by a velocity-dependent saturation function due to the DC motor's characteristics:
\begin{align}
\tau_{\text{sat}}^{+}(\dot{q}) &= \tau_0 \left(1 - \frac{|\dot{q}|}{\dot{q}_{\text{max}}}\right), \hspace{2mm}
\tau_{\text{sat}}^{-}(\dot{q}) &= -\tau_0 \left(1 - \frac{|\dot{q}|}{\dot{q}_{\text{max}}}\right),
\end{align}
where $\tau_0$ is the stall torque and $\dot{q}_{\text{max}}$ is the maximum no-load velocity. The final applied torque is given by:
\begin{equation}
\tau_{\text{applied}} = \eta \cdot \text{clip}\left( \tau_b,\ \tau_{\text{sat}}^{-}(\dot{q}),\ \tau_{\text{sat}}^{+}(\dot{q}) \right),
\end{equation}
where $\eta$ is a randomized factor that accounts for variations in motor torque constant and motor driver's efficiency.

All model parameters are resampled for each actuator at the beginning of every training episode. This comprehensive randomization strategy forces the control policy to adapt to a wide spectrum of actuator imperfections, significantly improving sim-to-real transfer performance.

\begin{figure}[t]
    \centering
    \includegraphics[width=0.9\linewidth]{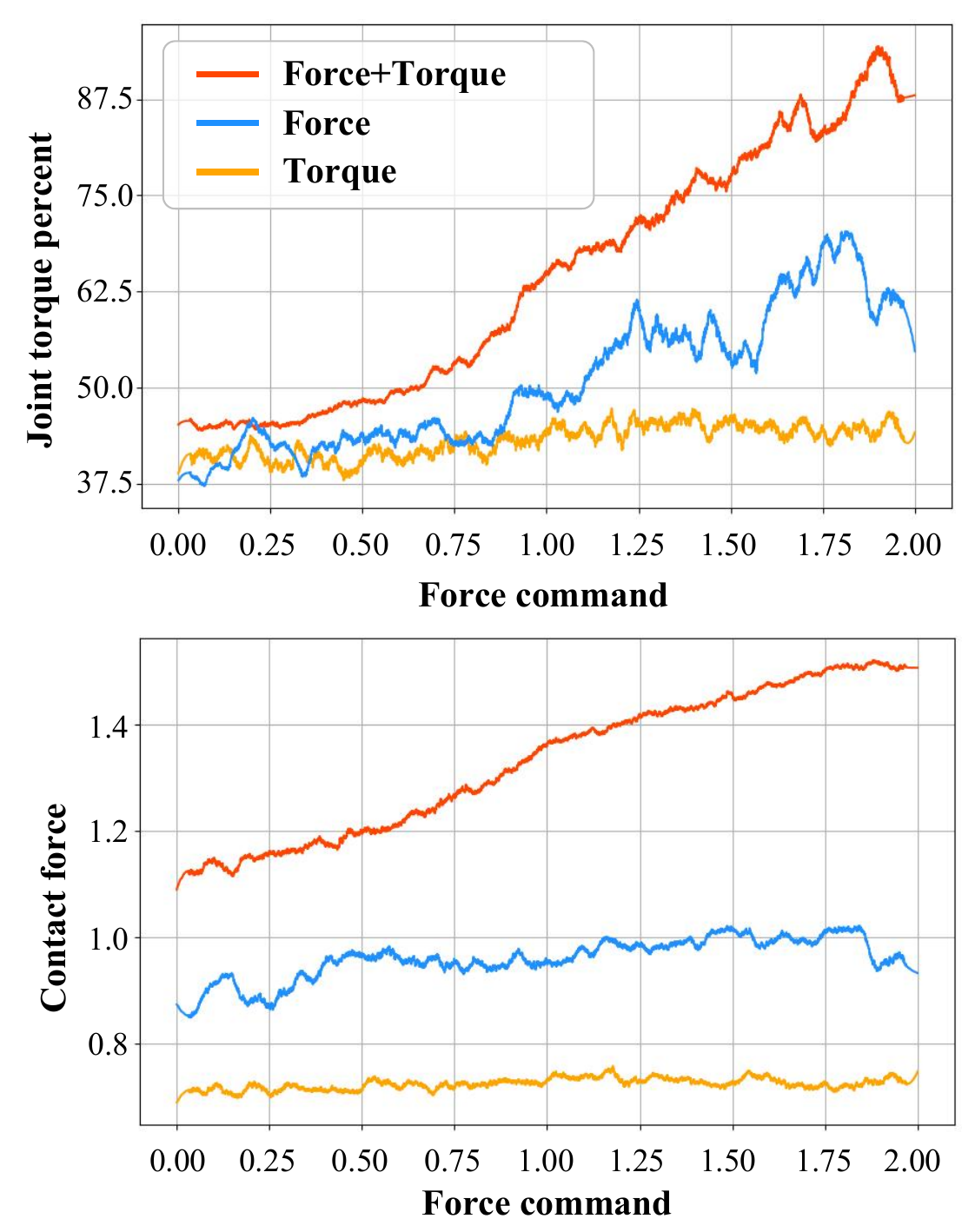}
    \caption{Varying joint torque and contact forces with force commands under different reward settings.}
    \label{fig:Force_command}
    \vspace{-5mm}
\end{figure}

\begin{table*}
\vspace{2mm}
\caption{Ablation analysis of multiple observation combinations in the real world.}
\vspace{-1mm}
\centering
\begin{tabular}{@{}cccc|ccccc@{}}
\toprule
\begin{tabular}[c]{@{}c@{}}Contact \\ center\end{tabular} & \begin{tabular}[c]{@{}c@{}}Contact \\ force\end{tabular} & \begin{tabular}[c]{@{}c@{}}Force\\ Weighted\end{tabular} & \begin{tabular}[c]{@{}c@{}}Orientation \\ representation\end{tabular} & Cons. success trials(sorted) & Average & Median & \begin{tabular}[c]{@{}c@{}}Success \\ time(Ave.)\end{tabular} & \begin{tabular}[c]{@{}c@{}}Time to fall\\ (Ave.)\end{tabular} \\ \midrule
\checkmark & \checkmark & \checkmark & 6D &8,10,12,15,19,25,26,35,46,55  &\textbf{25.1}  & \textbf{22.0} &3.36  & \textbf{84.3} \\
\checkmark & \checkmark & \checkmark & Quaternion & 1,1,1,1,2,3,4,5,5,6& 2.9 & 2.5 & 5.24 & 15.2 \\
\checkmark & \checkmark & × & 6D & 3,6,8,10,11,13,13,15,21,22 & 12.2 & 12.0 & 3.75 &45.7  \\
\checkmark & × & \checkmark & 6D & 2,6,10,11,13,14,16,18,21,21 & 13.2 & 13.5 & 3.19 & 42.1 \\
\checkmark & × & × & 6D &1,3,6,6,7,8,9,11,11,13  &7.5  & 7.5 &4.03  & 30.2 \\
× & × & × & 6D & 1,1,1,1,1,1,1,1,1,2 & 1.1 & 1.0 & 2.82 &3.10  \\
\checkmark & \checkmark & \checkmark & × &2,4,5,8,11,12,13,16,17,24  &11.2  & 11.5 &\textbf{2.59} & 29.0 \\ \bottomrule
\end{tabular}

\label{tab:abl_cube}
\vspace{-3mm}
\end{table*}

\begin{figure*}[t]
    \centering
    \includegraphics[width=0.95\linewidth]{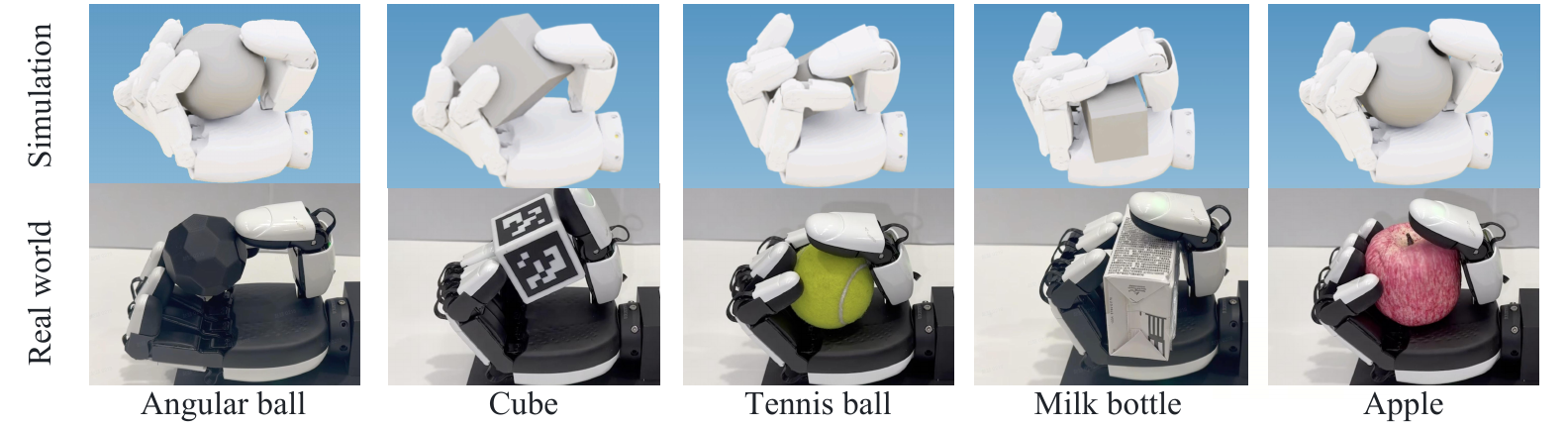}
    \caption{Visualization results of force-adaptive grasping tasks in Real-world and simulation environments}
    \label{fig:grasp}
    \vspace{-4mm}
\end{figure*}

\begin{figure*}[t]
    \centering
    \includegraphics[width=0.95\linewidth]{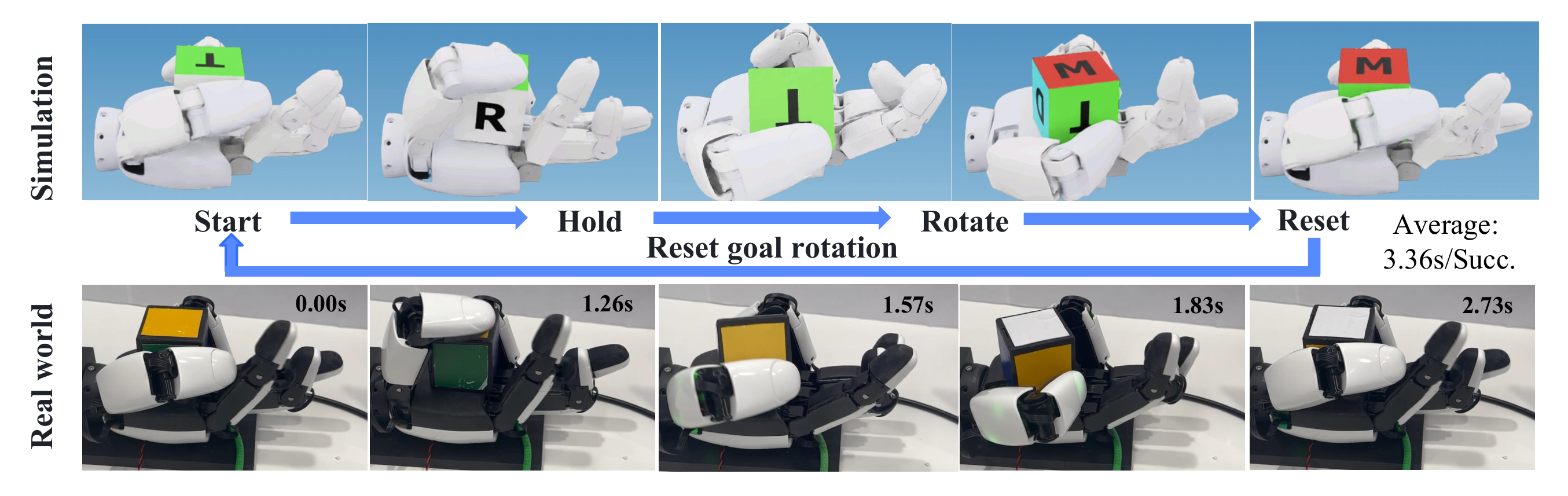}
    \vspace{-3mm}
    \caption{Visualization results of in-hand manipulation tasks in real-world and simulation environments}
    \label{fig:rot_vis}
    \vspace{-5mm}
\end{figure*}

\begin{figure}[!b]
\vspace{-5mm}
    \centering
    \includegraphics[width=1\linewidth]{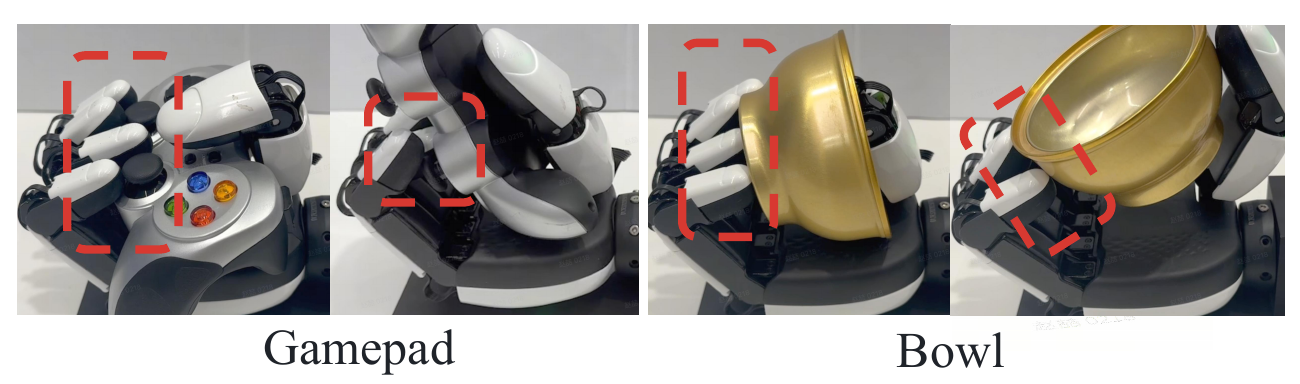}
    \caption{Force-adaptive grasping of irregularly shaped objects that are unseen during training.}
    \label{fig:irregularly}
\end{figure}

\section{Experimental Results}
\label{sec:exp}

Part of the simulation and experimental results are included in Section \ref{sec:method} (Fig. \ref{fig:contact_point}, \ref{fig:grasp_force_command}, \ref{fig:Force_command}) for providing evidences to support the methodologies with better logical flow, here we present more results on the following three aspects.  

\subsubsection{Force-adaptive grasping}We conducted an ablation study on the reward terms $R_{\text{Force}}$ and $R_{\text{torque}}$, which are designed to regulate the magnitude of the grasping force. As summarized in Tab.~\ref{tab:abl_force}, using both terms together leads to stronger and wider ranges of contact forces and joint torques, reflecting more forceful and dynamic grasping.
When only $R_{\text{Force}}$ or $R_{\text{torque}}$ is used, both the contact force and joint torque ranges are reduced, indicating less assertive grasping. These findings confirm that both reward components are necessary for achieving robust and adaptive grasping.

Further we investigated the relation of contact force and torque as functions of the force command, as shown in Fig.~\ref{fig:Force_command}. Both contact force and torque exhibit an approximately linear positive correlation with the commanded force intensity. It can be observed that the strength of correlation decreases in the following order: when both reward terms are used, when only 
$R_{\text{Force}}$ is used, and when only $R_{\text{torque}}$ is applied.

\subsubsection{In-Hand Object Rotation}

Real-world ablation experiments were conducted to assess each observation component. Policies were evaluated on ten trials using consecutive successes, average duration, and time to failure as metrics (Tab.~\ref{tab:abl_cube}). The full configuration(force-weighted contact center, contact force, and 6D orientation) achieved optimal performance with 25.1 consecutive successes and a 3.36-second average duration. Replacing 6D with quaternions caused a significant decrease to 2.9 successes, showing the need for a continuous rotation representation. Removing the force-weighted contact center reduced the successes to 12.2, while skipping the contact force resulted in 13.2 successes. Ablating both components further decreased performance to 7.5 successes. The baseline without contact sensing performed poorly (1.1 successes), confirming the importance of the tactile feedback. A model without explicit orientation encoding had 11.2 successes, underperforming that of the full configuration. These results show that \textit{contact information, force sensing, force weighting, and 6D rotation} are critical to robust in-hand rotation.

\subsubsection{Simulation and Experimental Results} As shown in Fig.~\ref{fig:grasp} and Fig.~\ref{fig:rot_vis}, we present simulation and real-world results for both force-adaptive grasping and in-hand rotation. The policies exhibit consistent behavior across domains. Fig.~\ref{fig:grasp_force_command} further demonstrates the response to different force commands, showing that the policy modulates grasp intensity as intended. Moreover, Fig.~\ref{fig:irregularly} illustrates successful grasping of unseen irregular objects, indicating that the policy generalizes to novel shapes with stable and conforming contact.

\begin{table}[]
\vspace{2mm}
\caption{Ablation on the rewards $R_{\text{Force}}$ and $R_{\text{torque}}$. }
\vspace{-1mm}
\centering
\begin{tabular}{@{}cc|cc@{}}
\toprule
$R_{\text{Force}}$      & $R_{\text{torque}}$     & Contact force range & Joint torque percent  \\ \midrule
\checkmark & \checkmark &  \numrange[range-phrase ={\,$\sim$\,}]{44.50}{93.92}                   &         \numrange[range-phrase ={\,$\sim$\,}]{1.09}{1.52}               \\
\checkmark & ×           & \numrange[range-phrase ={\,$\sim$\,}]{37.29}{70.02}                     &        \numrange[range-phrase ={\,$\sim$\,}]{0.69}{0.75}                    \\
          × & \checkmark &      \numrange[range-phrase ={\,$\sim$\,}]{38.45}{47.12}               &            \numrange[range-phrase ={\,$\sim$\,}]{0.85}{1.02}               \\ \bottomrule
\end{tabular}

\label{tab:abl_force}
\vspace{-7mm}
\end{table}

\section{Conclusion}

We propose a reinforcement learning framework and a suite of sim-to-real techniques designed to achieve robust, force-aware dexterous manipulation using a five-finger dexterous hand. To overcome the challenge of simulating tactile sensing without incurring high computational costs, we introduce a novel approximation method based on parallel distance computation. This approach achieves good contact approximation that enables zero-shot sim-to-real transfer, while ensuring the high sample efficiency required for training in simulation. To the best of our knowledge and experience, our simulation-based RL techniques are broadly applicable for training dexterous manipulation policies on a wide range of commercial humanoid hands, thereby facilitating research progress toward human-level dexterity via simulation-based learning paradigm.

\bibliographystyle{IEEEtran}
\bibliography{IEEEabrv, reference}

\end{document}